# Union-net: Lightweight deep neural network model suitable for small data sets

Jingyi Zhou [1], Qingfang He [2], Zhiying Lin [2]

[1] Institute of Scientific and Technical Research on Archives, Beijing, 100050, China
[2] Institute of Computer Technology, Beijing Union University, Beijing, 100101, China
Corresponding Author: Qingfang He
Email address: qingfang@buu.edu.cn

## Abstract

In recent years, deep learning has developed rapidly in the field of artificial intelligence and has made great achievements. Deep learning can achieve better performance with large data sets. In real applications, it is expensive to obtain big data. Therefore, research on deep learning for small data sets has been paid more and more attention. Judging from the published research results, the tasks of completing small data sets mostly use pre-trained models based on big data, which are realized through migration learning and fine-tuning technologies. There are two risks in using pre-trained models for commercial applications: One is the copyright issue of the pre-trained model or the data sample used by the pre-trained model. The other is that a suitable pre-training model needs to be selected for learning goals in a specific field. Therefore, this paper proposes a lightweight deep network model Union-net that adapts to small data sets. The structure of the model is as follows: Convolutional network units with the same input and different combinations constitute a union module. Each union module is equivalent to a multiple convolution layer (named union convolution in this paper). The serial input and output of three union modules form a  neural network. The results of the parallel output of the three union modules are merged and added, and the result of the addition is used as the input of the last layer of convolutional layer. So as to form a "shallow network" structure. The model can solve the problem of the traditional deep network model that the network is too deep and the conduction path is too long, which leads to the loss of the underlying information transmission. Based on the public data sets CIFAR10 and 17Flower, using multiple classic models and Union-net models, four small data set multi-class classification experiments and one large data set multi-class classification experiment were carried out respectively. The results show that the Union-net model is better than the classic model in the experiments on four small data sets. The code for this paper has been uploaded to https://github.com/yeaso/union-net .

## 1. Introduction

Training deep learning models often requires a lot of data, otherwise overfitting will occur.

The cost of making or collecting a large data set is very high (such as medical data sets. It is not only necessary to collect professional medical data samples, but also professional doctors to correctly label the samples.). Or it is impossible to obtain a big data (for example, when doing time series analysis, only records from a specific time period.). Therefore, in real applications, what can be obtained is generally a small data set. In recent years, the research of deep learning based on small data sets has been paid more and more attention. In order to solve the problem that deep learning models use small data sets to easily produce over-fitting, a variety of improved methods have been proposed. For example，on the basis of traditional enhancement data methods such as image rotation and distortion, the random non-repeated cropping method is used to balance the number of benign and malignant samples and expand the data set. Aiming at the samples that are difficult to cluster in the training set, the concept of "weak feature", "weak feature" sample extraction algorithm, adaptive adjustment, and secondary training algorithm are proposed to improve model training for small data sets[1]. In view of the small number of samples in the medical data set and the large difference in the number of samples in each category, the paper [2] uses model fusion, transfer learning, feature fusion technology, etc., to achieve good results in solving the multi-classification of breast cancer. Paper [3] proposed L1 and L2 regularization methods. Paper [4] proposed the dropout method. Literature [5] proposed an early stopping method. Paper [6] proposed methods to reduce the number of model parameters, enhance data, and generate new data with adversarial networks. Paper [7] proposed methods such as transfer learning and fine-tuning. Paper [8] proposed a 3D deep learning model AppendixNet. The model was pre-trained on a large number of YouTube videos called Kinetics, and then fine-tuned on a small dataset of 438 CT scans labeled for appendicitis. Paper [8] believes that pre-training the 3D model on a large natural video data set and then applying it to a specific small data set can improve the performance of the model.

    The above research is mainly based on the optimization of model hyperparameters, model pre-training, and fine-tuning techniques to improve the model's adaptability to small data sets. Most of the current research on deep learning applied to small data sets is focused on transfer learning (model pre-training and fine-tuning). That is, pre-training tasks with big data (such as the imageNet data set [9]), using the pre-trained model to apply to small data sets, to solve tasks similar to big data tasks on small data sets. At present, transfer learning has become a common method to realize deep learning based on small data sets [10, 11]. However, there are two major problems in practical applications when using the transfer learning method: First, the used big data may have copyright issues. For example, the samples in the big data set may come from the internet, and the license for use is not clear. There are also copyright laws implemented in many countries/regions that consider it illegal to use ImageNet pre-trained models for commercial applications. Secondly, pre-training models based on general large data sets (such as ImageNet) are not omnipotent. If the transfer learning scenario is very different between the source domain and the target domain, the results of machine learning are not ideal [12].

As we all know, given a large amount of data, even a simple learning model can solve complex tasks by memorizing [13, 14]. The hallmark of true artificial intelligence is that it can extract features from limited data. At present, in the field of deep learning, there are few studies on training learning models directly from small data sets instead of pre-training with other large data sets. Paper [15] carried out research on learning directly from small data sets and achieved good results. Paper [15] proposed to replace the commonly used cross-entropy loss function with the cosine loss function in the CNN model. Using this method, the performance of the model is improved on a data set with only a small number of samples in each category. Although the method proposed in paper [15] has achieved good experimental results, it still achieves the improvement of specific experiments by adjusting the conventional algorithm, which does not have universal applicability.

Our paper proposes a CNN model Union-net, which is a deep learning model with a "shallow network structure". This paper attempts to improve the model structure to achieve the model's adaptability to small data sets. The basic idea of the design: Convolutional network units with the same input and different combinations constitute a union module. Each union module is equivalent to a multiple convolution layer (named union convolution in this article). The serial input and output of three union modules form a neural network. The results of the parallel output of the three union modules are merged and added, and the result of the addition is used as the input of the last layer of convolutional layer. So as to form a "shallow network" network structure. Because the model structure is simple and has fewer model parameters, the problem of deep network model network is too deep, the conduction path is too long, and the underlying information transmission problem is solved. It solves the problem that deep network models are prone to overfitting in training small data sets.

The definition of a small data set is currently not clearly described, it is related to the completed tasks and the diversity of the data. According to the paper[16], in the psychiatric human research, samples less than 10,000 are "small" samples. Paper [8] uses 438 samples labeled with appendicitis as a small data set. Paper [15] uses about 20 to 100 samples per classification as a small data set. For image classification, a data set with less than 100 training samples per category is very small. Therefore, in the experiments of small data sets in this paper, the number of samples of each type is not more than 100 . All experiments in this paper are based on the public data set CIFAR10 [17] (reduced to less than 100 samples for each category) , 17 Category Flower Dataset (Called 17 Flower) [18] and etc..

Contributions of this paper: (i) A CNN model Union-net adapted to small data sets is proposed. (ii) The rationality of the model is deduced and verified. (iii) The concept of union convolution is proposed. (iv) The effectiveness of the model is verified by multiple experiments.

The rest of this article is arranged as follows: (i)Contents in Materials & Methods section: The proposed Union-net model, discussing the model structure, features, and union convolution; Experimental settings, discussing the datasets (CIFAR10, 17Flower) with all its details, and methodology stating how these were used in the experiments. (ii) Contents in Results and Discussion section: The experimental results were discussed and compared with other models.

(iii) Contents in Ablation Experiment section: Perform ablation experiments on the Union-net model to observe the influence of the combination of different structures in the Union module on the model. Finally, the research results are summarized, and the next research direction is proposed.

## 2. Materials & Methods
### 2.1 The Proposed Union-Net Model

The Union-net model is constructed based on convolutional neural network (CNN) technology [19]. The design inspiration of the model comes from three classic models: Resnet [20], Densenet [21], Xception [22]. The residual concept of Resnet is adopted. The Densenet's feature reuse idea is adopted. The channel merging method of multiple convolutions of the same input of Xception is changed to the information merging method of multiple different convolutions of the same input.

### 2.1.1 Union-net model structure

The structure of the Union-net model is shown in *Fig.1*. Due to the limitation of the map size, the batch standardization layer [23] and the activation function layer [24] are omitted in the *Fig.1*. For detailed configuration, please refer to the model structure diagram (The model structure diagram has been uploaded to https://github.com/yeaso/union-net ).

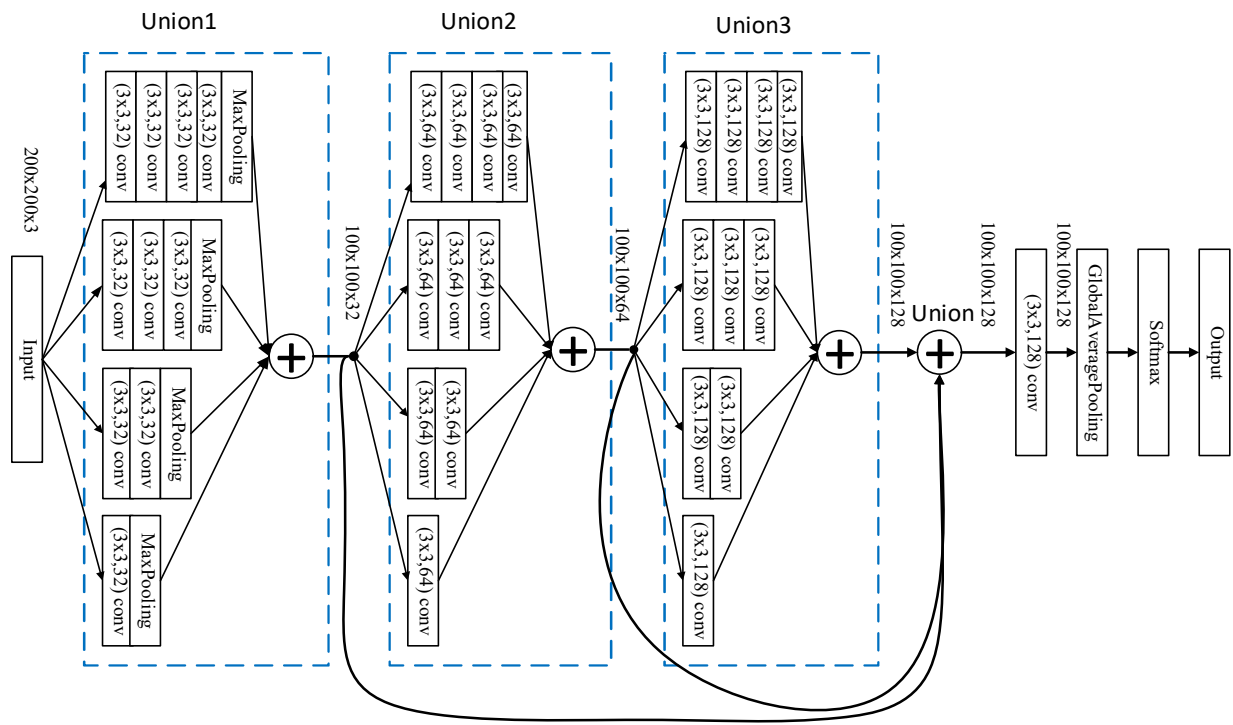

**Figure1. Union-net model structure.**

As can be seen from *Fig.1*, the model consists of an input layer, three Union modules, a 3x3x128 convolutional layer, a global average pooling layer [25] and a SoftMax classification output layer[26]. The "+" sign in each union module indicates the fusion and addition of the output results of the four groups of network units in the module. The fusion method used by Union-net is different from the fusion method used by the Xception [22] and Inception [27] models. Xception and Inception use channel splicing (called concatenate) [27]. Therefore, the shape of the feature map output by the Xception and Inception models is changed (that is, the number of channels increases). Union-net uses the addition of the information of multiple features. After the addition, the output shape of the feature does not change, and the number of channels remains unchanged. The output of Union1 is used as the input of Union2 module. The output of the Union2 module is used as the input of the Union3 module. The output of the Union3 module is merged and added with the output of Union1 and Union2, and the result of the addition is used as the input of the final convolutional layer of the model.

In order to further observe the feature output of the Union module in the model, a visualization tool is used to visualize the features of Union1, Union2, Union3, and the union output of the three nodes. Take the model trained in this paper based on the 17flower data set as an example to output visual features, as shown in *Fig.2*. It can be seen that from the low-end to the high-end of the model, the output feature map gradually changes from a detailed expression to an abstract expression.

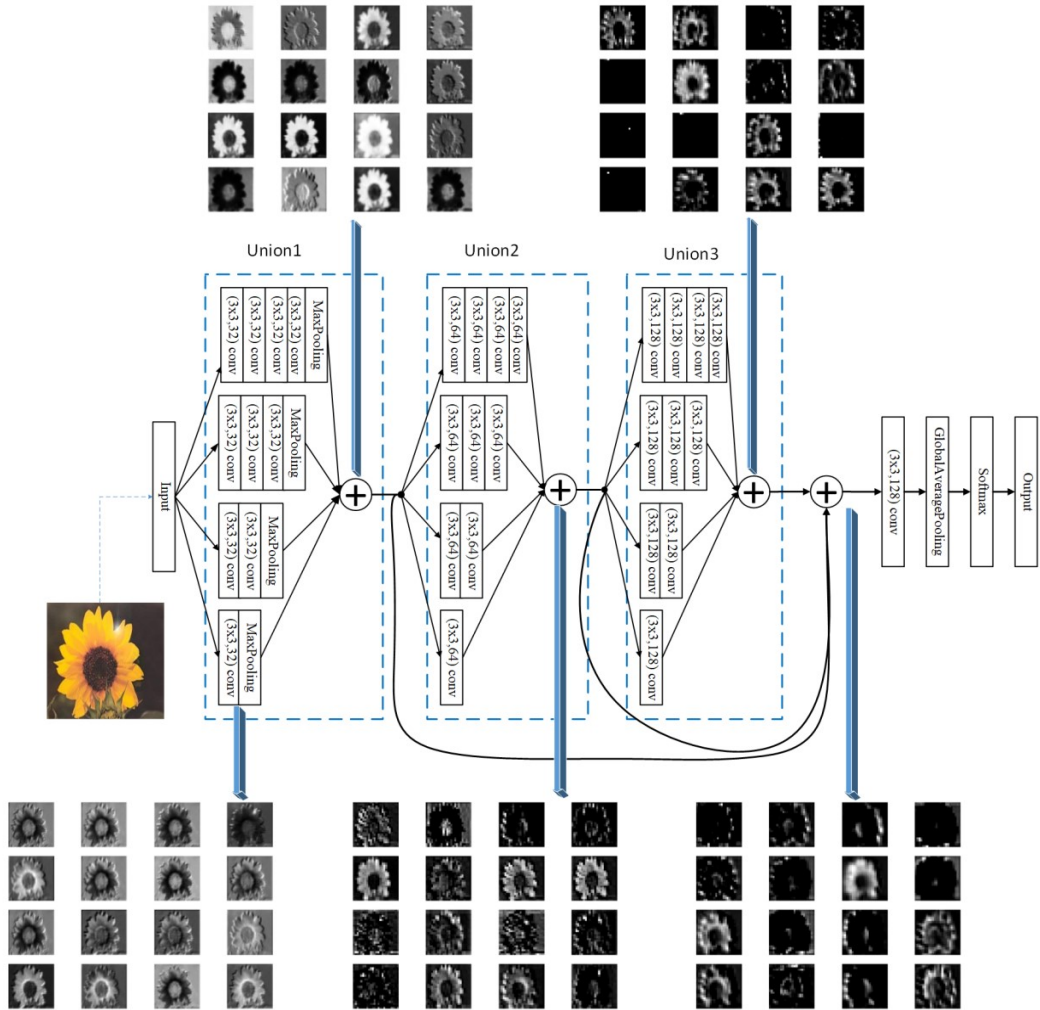

**Figure2. Inter-layer feature visualization of Union-net model.**

### 2.1.2 Union-net features

Judging from the model structure in *Fig.1*, Union-net has the following characteristics:
***I***. From the appearance, Union-net is a simple network with a simple structure and a shallow network structure. There are three modules containing multiple convolutions and one convolutional layer (Union1, Union2, Union3, and ((3x3,128)conv).
***II***. All convolution layers use 3x3 small convolution kernels to reduce model parameters and computational complexity. This model has a "shallow network" structure and uses a batch standardization layer [23], so the overfitting problem is solved. The dropout [4] used by the deep neural network model is not used here to prevent overfitting, because Union-net faces fewer data samples. If dropout is used, the acquired information is randomly discarded during the training process, which will affect the model's ability to fully capture sample features.

***III***. Union-net uses Maxpooling layer in the first union module [28]. According to related theories [28], The Maxpooling can reduce the deviation of the estimated mean value caused by the parameter error of the convolutional layer, and retain more texture information. The Maxpooling can retain more effective features, reduce dimensions, reduce parameters, and remove noise. These advantages are very beneficial for extracting features and suppressing overfitting. Therefore, Union-net uses the Maxpooling layer in the Union1 module. After the initial input data of the model is processed by Union1's internal convolution structure, the output features are pooled and filtered to provide more effective data for subsequent processing. In the final stage of the model, the Maxpooling layer is not used. The main consideration is that the model structure is shallow, and increasing the pooling layer has the risk of losing local information. Instead, the traditional fully connected layer (dense layer) is replaced by global average pooling. This method comes from references [29,30]. The dense layer parameters of the traditional convolutional neural network are huge. Replace dense with global average pooling, which greatly reduces the amount of parameters. This is helpful to suppress the overfitting of the model when training on a small data set.

***IV.*** The power of unity. Figure 3 shows the Union module structure. There are four groups of independent parallel convolutional neural network structures in the module. They are: 3x3 convolution kernel completes one layer of convolution (denoted as a); 3x3 convolution kernel completes two-layer stacked convolution (denoted as b); 3x3 convolution kernel completes three-layer stacked convolution (denoted as c); The 3x3 convolution kernel completes four-layer stacked convolution (denoted as d). The input of each group of convolution is the same, but each group has different receptive fields for the input features. Group a is the receptive field of the 3x3 convolution kernel. Group b is equivalent to the receptive field of the 5x5 convolution kernel. Group c is equivalent to the receptive field of the 7x7 convolution kernel. Group d is equivalent to the receptive field of the 9x9 convolution kernel [31]. Although the groups a, b, c, and d in the union module are simple convolutional neural network structures, they observe and extract features from different fields of view on the input features (shapes). The feature information obtained by them is added and merged, and the merged information is sent to the next union module. In the union module, the four groups of neural network units work individually, and the outputs of each group are combined together. From the perspective of the entire model, the outputs of the three union modules are combined together. This is also the origin of the model name Union-net.

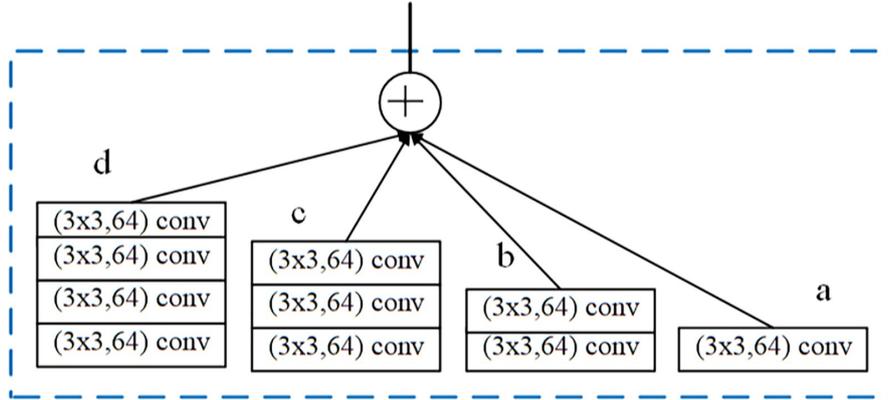

**Figure3. Union module structure.**

*V.* The number of convolution kernels (also called the number of channels) of all convolution layers in Union-net is less than the number of convolution kernels of traditional neural networks, and the maximum here is 128. The maximum number of channels of the traditional neural network convolution kernel is generally 1024 or even larger. Traditional neural networks hope to increase the number of channels to enhance the model's ability to identify features and feature extraction. This model has been able to recognize multiple features through the convolutional composite of multiple different fields of view in the union module, and also solves the problem of partial feature information loss due to the network being too deep. Therefore, Union-net uses a smaller number of channels, which does not affect the model's ability to extract features, but also greatly reduces model parameters and model complexity.

*VI.* Compared with the classic deep learning model, the Union-net model has fewer parameters and less floating point operations (FLOPs). *Table 1* shows the comparison between the classic network model and the Union-net model in terms of model parameters, model FLOPs. It can be seen from the *Table 1* that the parameters of Union-net is much smaller than most classic networks, and is similar to the light-weight networks MobileNetV2 [30] and EfficientNet-B0[33], and the FLOPs value of Union-net is much smaller than other models.

**Table 1 Comparison of features between classic network and Union-net model.**

| Model | Parameters | FLOPs |
|---|---|---|
| Union-net | 1.72M | 0.003B |
| VGG16[31] | 138M | 15.5B |
| Xception[22] | 23M | 8.4B |
| ResNet50[20] | 26M | 3.9B |
| InceptionResNetV2[32] | 56M | 13B |
| DenseNet169[21] | 14M | 3.5B |
| MobileNetV2 | 3.4M | 0.6B |
| EfficientNet-B0[33] | 5.3M | 0.39B |

### 2.1.3 Union convolution

This paper believes that the Union-net model can be regarded as a simple network structure. By analyzing the input and output among Union-net modules, this view is reasonable. In *Fig.3*, there are four groups of convolutional neural network structural units a, b, c, and d, and each group has the same input X. a is a neural network structural unit of a convolutional layer, and its output $Y_a$ is expressed by formula (1). b is a neural network structural unit of 2 convolutional layers, and its output $Y_b$ is expressed by formula (2). c is a neural network structural unit of 3 convolutional layers, and its output $Y_c$ is expressed by formula (3). d is a neural network structural unit of 4 convolutional layers, and its output $Y_d$ is expressed by formula (4). Here $w$ is the weight corresponding to the input $X$, $b$ is the bias parameter, $\sigma$ is the activation function.

$$Y_a = \sigma_{a1}(w_{a1} * X + b_{a1}) \tag{1}$$

$$Y_b = \sigma_{b2}(w_{b2} * (\sigma_{b1}(w_{b1} * X + b_{b1})) + b_{b2}) \tag{2}$$

$$Y_c = \sigma_{c3}(w_{c3} * (\sigma_{c2}(w_{c2} * (\sigma_{c1}(w_{c1} * X + b_{c1})) + b_{c2})) + b_{c3}) \tag{3}$$

$$Y_d = \sigma_{d4}(w_{d4} * (\sigma_{d3}(w_{d3} * (\sigma_{d2}(w_{d2} * (\sigma_{d1}(w_{d1} * X + b_{d1})) + b_{d2})) + b_{d3})) + b_{d4}) \tag{4}$$

Formula (5) is a general expression of formulas (1)-(4). Where $X_{i-1}$ represents the input of the previous layer of convolution, $i$ represents the four groups of convolutional neural networks a, b, c, and d, and $j$ represents the convolutional layer corresponding to each group of convolutional neural networks..

$$Y_i = \sigma_{ij}(W_{ij} * X_{i-1} + b_{ij}) \tag{5}$$

It can be seen that each union module has only one input and one output. The working process of the standard neural network convolutional layer: accepts an input, and obtains an output through convolution kernel processing. The union module accepts one input. The four groups of convolutional neural network structural units of the union module perform "complex convolution processing". The results of each group's processing are added together as the output of the union module. Therefore, this paper regards each union module as a union convolution unit, which is called union convolution. According to this idea, the model structure of *Fig.1* is simplified as shown in *Fig.4*. Obviously, Union-net is a "shallow" and simple deep neural network.

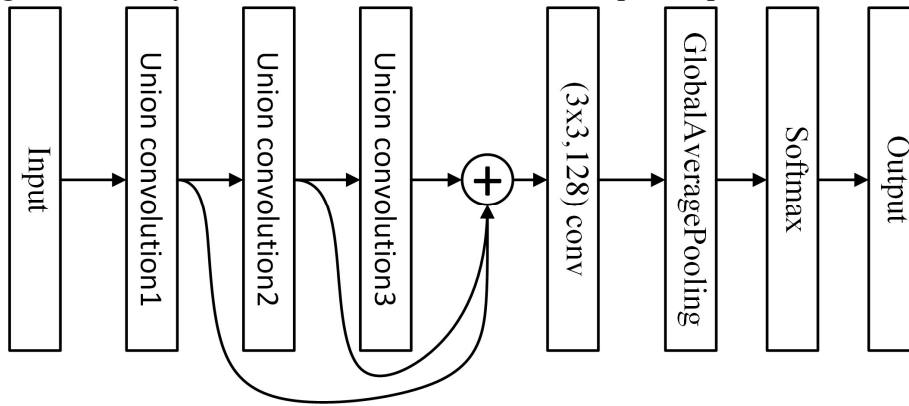

**Figure4. Union-net model simplified with union convolution.**

## 2.2 Experimental Settings

This paper uses CIFAR-10 data set [17], three small data sets made with the CIFAR-10 data set, and the 17Flower data set [18] to test the Union-net model.

### 2.2.1 Experiments based on CIFAR-10 data set and its small data sets

CIFAR-10 dataset was collected by Alex Krizhevsky, Vinod Nair, and Geoffrey Hinton[17].The CIFAR-10 dataset consists of 60000 (32x32) colour images in 10 categories, with 6000 images per category. There are 50000 training images and 10000 test images. Excluding the test set, each category has 5000 sample images, which is obviously not a small data set. Therefore, adjust the CIFAR-10 sample data to a small data set. In order to verify the generality of the model, from the CIFAR-10 dataset, randomly and without repetition, 30, 60, and 100 image samples of each category were selected to make three small data sets. The small data sets produced are denoted as CIFAR10_30, CIFAR10_60, and CIFAR10_100. For example, in the CIFAR10_30 data set, each category has 30 samples, and 10 categories have a total of 300 samples.

Experimental strategy: (i) Train the model separately based on the small data sets of CIFAR10_30, CIFAR10_60, and CIFAR10_100; Train the model based on the large data set of CIFAR-10 (5000 sample images for each category). In this way, under the same conditions, observe the performance of the model based on a large data set and the performance of the model based on a small data set. (ii) Define 10 classification labels {0,1, 2, 3, 4, 5, 6, 7, 8, 9} corresponding to {'airplane','automobile','bird','cat','deer ','dog','frog','horse','ship','truck' }. (iii) Use the CIFAR-10 test set to test and evaluate the trained models.

The main hyperparameters used in the experiment (The parameter values used in this paper are selected by the author based on our own experiences in the experiment, and are for readers' reference only.):
batch_size = 32
epochs = 38
Optimizers use Nadam. Nadam parameter settings (lr=0.01, beta_1=0.5, beta_2=0.999, epsilon=1e-08, schedule_decay=0.004).
Learning rate adjustment strategy: monitor='val_accuracy', factor=0.5, patience=2, verbose=1, mode='min', epsilon=0.0001, cooldown=0, min_lr=0.

### 2.2.2 Experiments based on 17 Flower dataset

17 Flower dataset is collected and produced by the Visual Geometry Group of Oxford University. This data set selects 17 kinds of flowers that are more common in the United Kingdom, each of which has 80 color pictures of different sizes, and the entire data set has 1360 pictures. The images in each category have great posture and light changes, and the images between different categories are also very similar.

Because all categories are very similar (that is, the types of flowers), and all flowers have many common structures (such as petals, stamens, pistils, etc.), the 17 Flower classification is a fine-grained classification task. As we all know, fine-grained classification tasks are the most challenging. Excluding the test set and validation set, there are only about 60 samples for each category that can participate in training. With so few training samples, challenging this task becomes more interesting.

Experimental strategy: Define 17 classification labels {0,1, …, 16 }, corresponding to 17 kinds of flowers {'Buttercup','Colts'foot', ...,' Cowslip' }.
The data set is divided into training set, validation set and test set, with an allocation ratio of 8:1:1.

The data set provides very few samples, with 64 training samples for each type, 8 verification samples for each type and 8 test samples for each type. In order to make full use of the sample and improve the generalization ability of the model, this paper adopts the 10-fold cross-validation method. Each sample is verified once and tested once without repeating it. The model is trained and tested for 10 times. Average the results of 10 experiments as the final evaluation result of the model. The specific allocation of 10-fold cross-validation is shown in *Fig.5*. The samples in each category are randomly divided into 10 equal parts, each with 8 samples, denoted as D0, D 1, D2, D3, D 4, D5, D6, D7, D8, D9. Take D0 for testing, D1 for verification, and others for training. Take D1 for testing, D2 for verification, and others for training. In this order, until D9 is used for testing and D0 is used for verification.

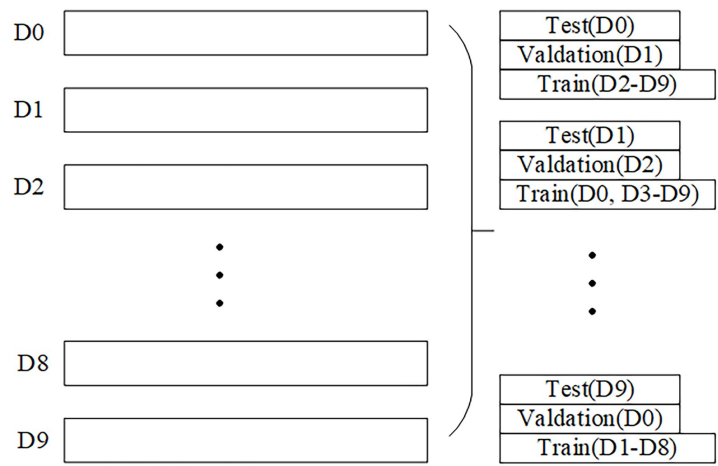

**Figure5. The specific scheme of 10-fold cross-validation.**

The main hyperparameters used in the experiment:
epochs = 38
batch_size = 32
Optimizers use Nadam. Nadam parameter settings (lr=0.01, beta_1=0.9, beta_2=0.999, epsilon=1e-08, schedule_decay=0.004).
Learning rate adjustment strategy: monitor='val_accuracy', factor=0.5, patience=2, verbose=1, mode='min', epsilon=0.0001, cooldown=0, min_lr=0.

## 3. Results and Discussion

The model experiment process record curve based on cifar10_30, cifar10_60, cifar10_100 small data set and cifar10 data set is shown in *Fig.6*. In the legend, Training acc means training accuracy, and Validation acc means verification accuracy.

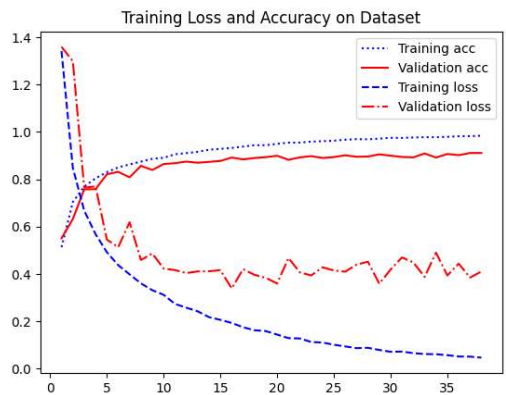

a. Based on the cifar10 data set

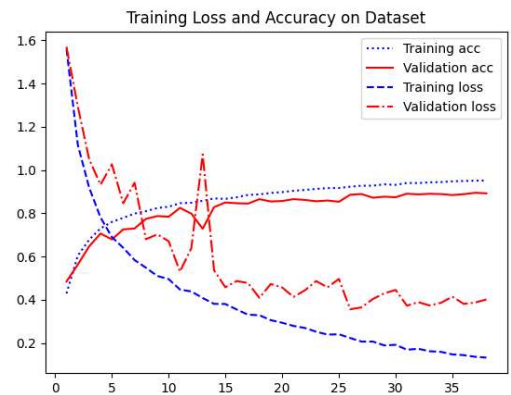

b. Based on the cifar10_30 data set

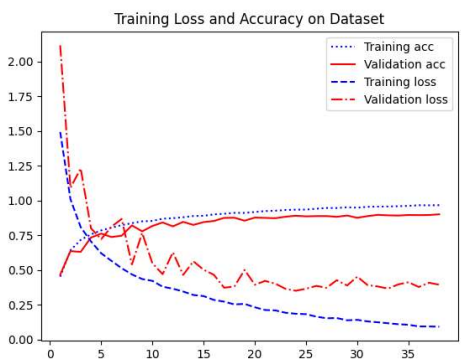

c. Based on the cifar10_60 data set

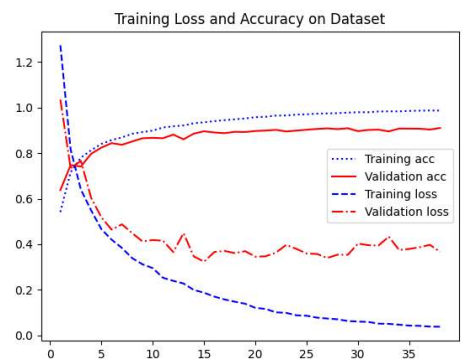

d. Based on the cifar10_100 data set

**Figure6. Union-net experimental record curve based on cifar10 data sets.**

This set of curves records the changes in the training accuracy, verification accuracy, training loss, and verification loss of the model during 38 epochs. Observing the four sets of curves, the model training process is generally stable. The performance of the model based on cifar10 large data samples is slightly better than the performance of the other three groups based on small data sets. Although the performance of the three groups of models based on small data sets is slightly lower than that based on large data sets, the model's ability to adapt to small data sets still has a better performance.

The experimental results of Union-net based on CIFAR10_30, CIFAR10_60, CIFAR10_100 and CIFAR10 data sets are shown in *Table 2*.

**Table 2 Union-net accuracy based on CIFAR10_30, CIFAR10_60, CIFAR10_100 and**

**CIFAR10 data sets.**

| CIFAR10_30 | CIFAR10_60 | CIFAR10_100 | CIFAR10 |
|---|---|---|---|
| 0.8988 | 0.9044 | 0.9068 | 0.9179 |

Experiments show that when the number of training samples for each type is 30, 60, 100, 5000, the classification ability of the model increases as the number of samples increases, and the classification accuracy increases. Although the large data set (cifar10) with 5000 training samples of each category performs best, the difference is not significant compared with the results of the three small data sets.

Based on the 17 Flower small data set, the experimental recorded curve of the first fold of the Union-net model is shown in *Fig.7*. This curve recorded the changes of the model's training accuracy, verification accuracy, training loss, and verification loss in 38 Epochs. The curve shows that the model training process has less fluctuation and the model converges faster.

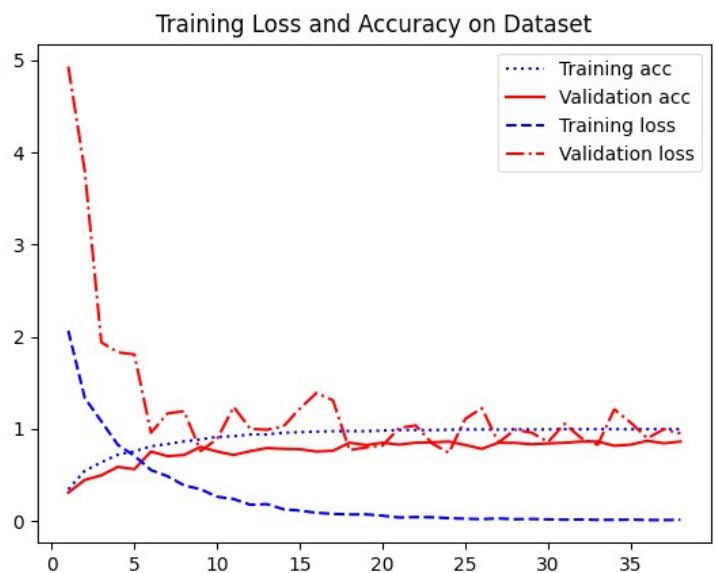

**Figure7. The experimental recorded curve of the Union-net model based on the first fold 17Flower data set.**

Based on the 17 Flower small data set, the Union-net model performs a 10-fold cross-validation experiment. The average recognition rate of the model reached 87%. In order to verify the generality of the design model, based on the CIFAR10_30, CIFAR10_60, CIFAR10_100, CIFAR10 and 17Flower data sets, using the same server and using basically the same hyperparameters, experiments were performed on CNN[34], DenseNet[35], Vgg16, Resnet50, MobileNetV2, Efficientnet[33], BReG-NeXt[36], ASMNet[37]models respectively. The comparison between these experimental results and the results of Union-net is shown in *Table 3*.

**Table 3 Comparison of experimental results between Union-net and other models.**

| Data set | Model | Accuracy |
|---|---|---|

| Dataset | Model | Accuracy |
|---|---|---|
| CIFAR10_30 | Union-net(ours) | 89.88% |
| | CNN[34] | 55.60% |
| | DenseNet[35] | 75.94% |
| | Vgg16 | 78.99% |
| | Resnet50 | 82.88% |
| | MobileNetV2 | 85.49% |
| | Efficientnet[33] | 84.62% |
| | BReG-NeXt[36] | 79.55% |
| | ASMNet[37] | 83.34% |
| CIFAR10_60 | Union-net(ours) | 90.44% |
| | CNN | 64.55% |
| | DenseNet | 79.99% |
| | Vgg16 | 84.17% |
| | Resnet50 | 85.96% |
| | MobileNetV2 | 84.36% |
| | Efficientnet | 86.69% |
| | BReG-NeXt | 80.79% |
| | ASMNet | 84.84% |
| CIFAR10_100 | Union-net(ours) | 90.68% |
| | CNN | 68.97% |
| | DenseNet | 85.88% |
| | Vgg16 | 86.64% |
| | Resnet50 | 87.71% |
| | MobileNetV2 | 86.56% |
| | Efficientnet | 87.22% |
| | BReG-NeXt | 81.57% |
| | ASMNet | 85.06% |
| CIFAR10 | Union-net(ours) | 91.79% |
| | CNN | 69.65% |
| | DenseNet | 90.79% |
| | Vgg16 | 90.78% |
| | Resnet50 | 92.75% |
| | MobileNetV2 | 91.98% |
| | Efficientnet | 94.19% |
| | BReG-NeXt | 91.52% |
| | ASMNet | 92.38% |
| 17Flower | Union-net(ours) | 86.58% |
| | CNN | 53.66% |

|   | DenseNet | 72.48% |
|---|----------|--------|
|   | Vgg16 | 80.85% |
|   | Resnet50 | 81.66% |
|   | MobileNetV2 | 82.74% |
|   | Efficientnet | 84.59% |
|   | BReG-NeXt | 80.63% |
|   | ASMNet | 83.11% |

There are 5000 samples for each category involved in training in the CIFAR10 data set. CIFAR10_30, CIFAR10_60, CIFAR10_100, and 17Flower datasets have 30, 60, 100, and 64 samples for each category participating in the training. From the experimental results, Union-net performed outstandingly on the small data sets of CIFAR10_30, CIFAR10_60, CIFAR10_100, and 17Flower. Based on the experiment of CIFAR10 large data set, except for CNN [33], the performance of each model has little difference. This result shows that the model Union-net proposed in this paper can better adapt to small data sets and have high model performance.

## 4. Ablation Experiment

In order to further verify whether the structural combination of the Union modules of the Union model is reasonable, based on the 17flower data set, split experiments and expansion experiments were performed on the union modules. As can be seen from the structure of the Union module in Figure 3, the Union module is composed of four groups of convolutional neural network units a, b, c, and d, so it is divided into the following experimental groups:
$U_a$, $U_b$, $U_c$, $U_d$, $U_{ab}$, $U_{ac}$, $U_{ad}$, $U_{bc}$, $U_{bd}$, $U_{cd}$, $U_{abc}$, $U_{abd}$, $U_{acd}$, $U_{bcd}$.
The expansion experiment is to expand the four groups of convolutional neural network units a, b, c, and d of the Union module into five groups of convolutional neural network units a, b, c, d, and e, where e is a 3x3 convolution kernel Five-layer stacked convolution (denoted as e), this experimental group is denoted as $U_{abcde}$.
Note: $U_a$ means that the Union module is composed of one group of convolutional neural network units a; $U_{abc}$ means that the Union module is composed of three groups of convolutional neural network units a, b, and c. The names of other experimental groups indicate similar meanings.
The parameter settings are exactly the same as those in the experiment in section 2.2.2, and the above 15 experimental groups were tested with the same computer. The record curve of each group of results obtained in the experiment is shown in *Fig.8*.

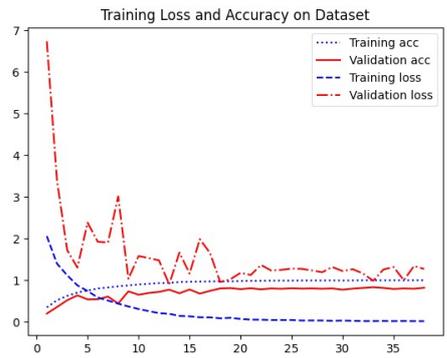

**a.     $U_a$**

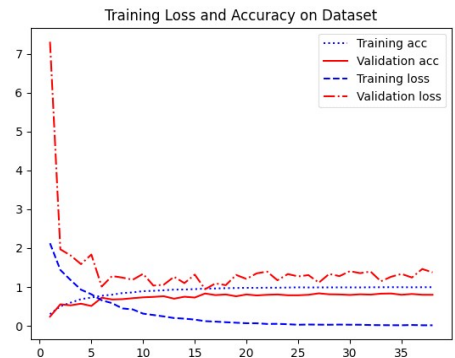

**b.     $U_b$**

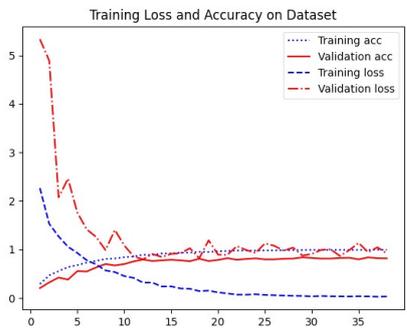

**c.     $U_c$**

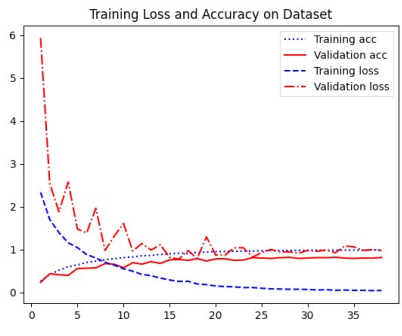

**d.     $U_d$**

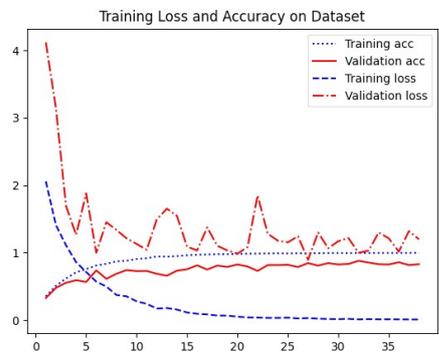

**e.     $U_{ab}$**

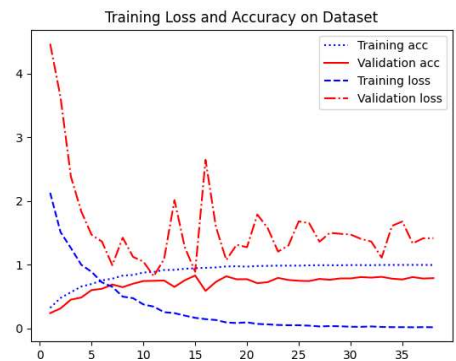

**f.     $U_{ac}$**

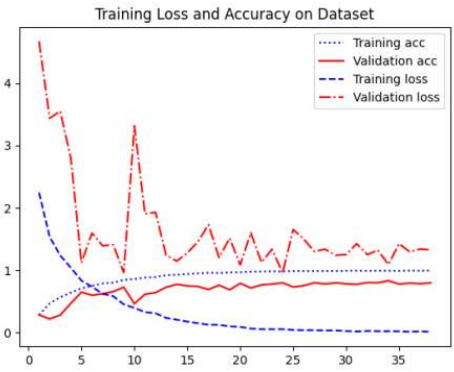

**g.      U_{ad}**

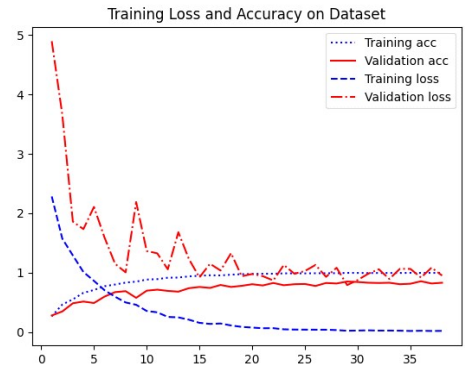

**h.      U_{bc}**

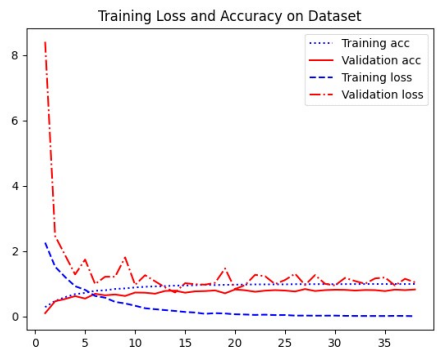

**i.  U_{bd}**

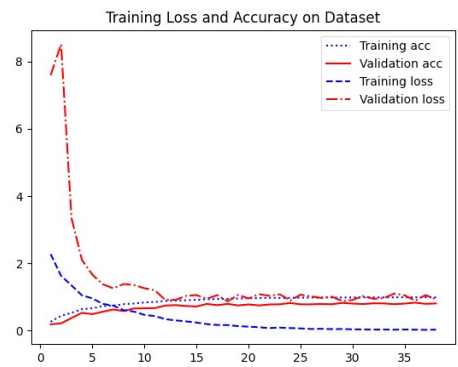

**j. U_{cd}**

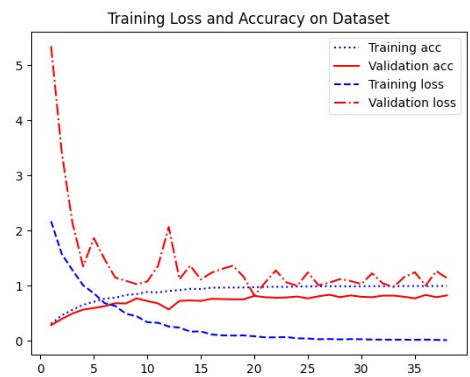

**k.  U_{abc}**

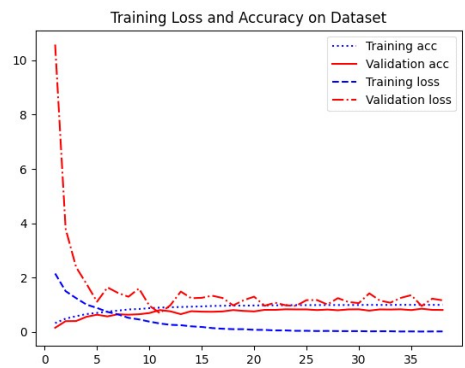

**l. U_{abd}**

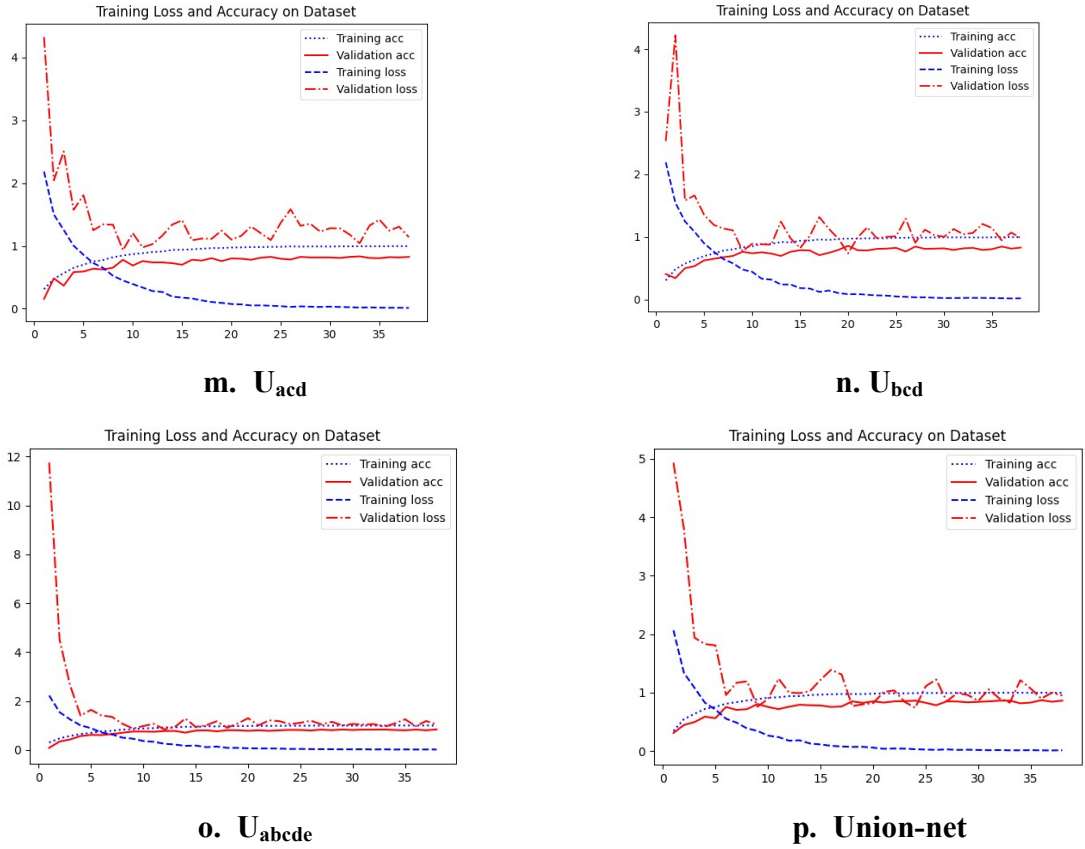

**Figure 8: Split experiment, extended experiment and Union-net model experiment record curve based on the 17flower data set.**

The model experiment results of 15 experimental groups are shown in *Table 4*.

**Table 4 Comparison table of split experiment, expanded experiment results and Union model results based on the 17flower data set.**

| $U_a$ | $U_b$ | $U_c$ | $U_d$ | $U_{ab}$ | $U_{ac}$ | $U_{ad}$ | $U_{bc}$ | $U_{bd}$ | $U_{cd}$ | $U_{abc}$ | $U_{abd}$ | $U_{acd}$ | $U_{bcd}$ | $U_{abcde}$ | Union |
|---|---|---|---|---|---|---|---|---|---|---|---|---|---|---|---|
| 0.8015 | 0.8051 | 0.8125 | 0.8088 | 0.8235 | 0.7794 | 0.7904 | 0.8199 | 0.8235 | 0.8051 | 0.8125 | 0.8125 | 0.8162 | 0.8235 | 0.8665 | 0.8658 |

In order to observe the difference between the results of the ablation experiment and the experiment of the Union-net model, the experimental results of each group and the experiment result of the Union-net model are represented by a line graph, as shown in *Fig.9*.

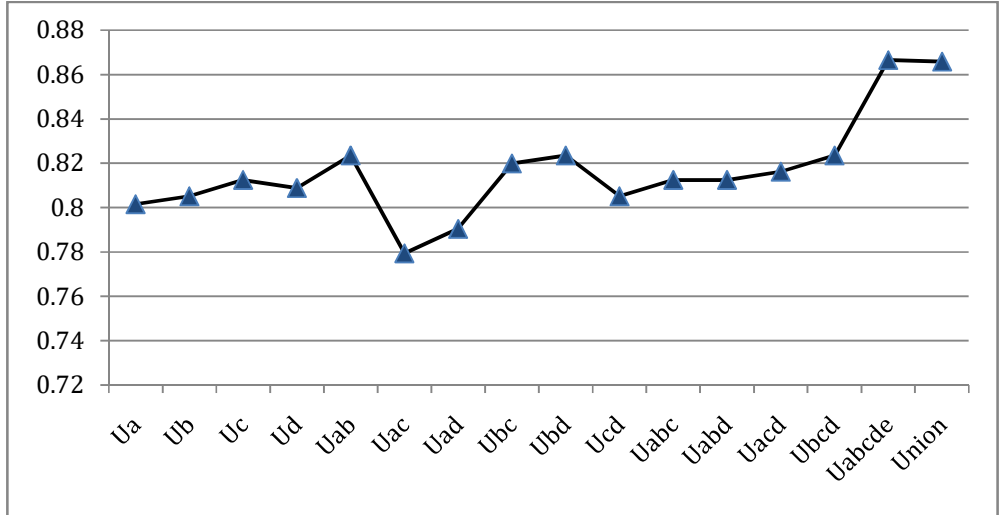

Figure 9 Analysis curve of split experiment and extended experiment results based on 17flower data set

It can be seen from *Fig.8* and *Fig.9* that except for the $U_{ac}$ and $U_{ad}$ groups, the differences between the other split experimental groups are not very large. The performance of the two groups of $U_{ac}$ and $U_{ad}$ lags behind $U_a$, $U_c$, $U_d$, indicating that $U_a$ combined with $U_c$ or with $U_d$, their combined results lag behind $U_a$, $U_c$, $U_d$. $U_{ab}$ is better than $U_a$ and $U_b$. $U_{bc}$ is better than $U_b$ and $U_c$. $U_{bd}$ is better than $U_b$ and $U_d$. The combination of $U_{cd}$ lags behind $U_c$ and $U_d$. From the results, without the participation of one-layer convolution a or two-layer convolution b, the effect of the combination is reduced. The results of $U_{abc}$, $U_{abd}$, $U_{acd}$, and $U_{bcd}$ show that the combination of any three of a, b, c, and d becomes stable, and the combined result no longer lags behind the single $U_a$, $U_b$, $U_c$, and $U_d$. $U_{abcde}$ and Union-net (ie $U_{abcd}$) bring out the advantages of the combination. Their performance is significantly better than the results of each split group. Although $U_{abcde}$ has one more e than Union-net, the experimental results are not significantly improved compared to Union-net. The structure of $U_{abcde}$ model is much more complicated than Union-net, and the convolution calculation amount of each Union module of $U_{abcde}$ has changed from the original a+b+c+d=1+2+3+4=10 to a+b+c+d+e=1+2+3+ 4+5=15. Therefore, this paper selects $U_{abcd}$ structure as the designed model.

## 5. Further verification

In order to verify the applicability of the Union-net model, experiments on the model are based on MNIST[38], Fashion-MNIST[39], MiniImageNet[40], and CUB200_2011 data set [41]. The method of making the training set is as follows: (i) 10 categories of training samples of MNIST, each of which randomly selects 100 samples from the original data set as training samples. (ii) 10 categories of training samples of Fashion-MNIST, each of which randomly selects 100 samples from the original data set as training samples. (iii) MiniImageNet contains 60,000 color images in 100 categories, each of which has 600 samples. Randomly select 10 categories from its 100 categories, each category has 100 training samples, and a total of three data sets are obtained, denoted as MIN (R1), MIN (R2), and MIN (R3). (iv) The CUB200_2011 data set has a total of 200 categories, each with about 60 samples, and a total of 11788 pictures. It is a fine-grained classification data set about birds. 10 categories were randomly selected from its 200 categories, and a total of three data sets were obtained, denoted as CUB (R1), CUB (R2), and CUB (R3). The experimental equipment and parameter configuration are consistent with the aforementioned ablation experiment. The measured experimental results are shown in *Table 5*.

**Table 5 Test results of Union-net model on various data sets.**

| Data set | MNIST | Fashion-MNIST | MIN (R1) | MIN (R2) | MIN (R3) | CUB (R1) | CUB (R2) | CUB (R3) |
|---|---|---|---|---|---|---|---|---|
| Accuracy | 0.9867 | 0.9888 | 0.9357 | 0.9204 | 0.9401 | 0.8924 | 0.9033 | 0.8896 |

Judging from the test results in *Table 5*, the Union-net model performs well overall. The model needs further improvement in fine-grained classification.

## 6. Conclusions and Future works

In view of the complex structure of the current deep neural network model, the huge model parameters, the need for larger data sets, the need for more advanced equipment for training, the complex parameter adjustment of the model, and the time-consuming and labor-intensive training characteristics, this paper proposes the idea of union convolution. The Union-net model built based on union convolution has a simple structure, few parameters, strong feature extraction ability, and good adaptability to small data sets.The above-mentioned problems of deep learning are solved from the model structure. Experiments show that the Union-net model proposed in this paper can perform well on classification problems based on small data sets. Union-net has high value in promoting the widespread application of deep learning. The Union-net model performs better on less-class classification of small data sets (in this paper, 10-class classification and 17-class classification). This article does not conduct further research on more-class classification (such as 100-class classification, 1000-class classification), which are challenges in the future. The source code of this paper has been released, and we hope peer experts can give us some suggestions and discover problems with the model to further improve the performance of the model.

## Funding

This work was supported by  Academic Research Projects of Beijing Union University (No.XP202021).

## Conflicts of interest/Competing interests

The authors declare no competing financial interests.

## Code or data availability

The source code of the program supporting the results of this study has been uploaded to github. The URL is https://github.com/yeaso/union-net

## Authors' Contributions

Jingyi Zhou: Software, Validation,Data curati.

Qingfang He: Writing- Original draft preparation, Software，Writing- Reviewing and Editing,Conceptualization, Methodology.

Zhiying Lin: Visualization, Investigation.

## Ethics approval

Not applicable.

## Consent to participate

Not applicable.

## Consent for publication

Not applicable.

## References

1. Qingfang He, Hui Wang, Guang Cheng. Research on Classification of Breast Cancer Pathological Tissues with Adaptive Small Data Set. Computer Science, 2021, 48(6A): 67-73.
2. Qingfang He, Guang Cheng, Huimin Ju. BCDnet:Parallel heterogeneous eight-class classification model of breast pathology. PloS one vol. 16,7 e0253764. 12 Jul. 2021, doi:10.1371/journal.pone.0253764.
3. Wu Guoning, Hu Huifeng, and Yu Mengmeng. Research on Regularization Methods in Deep Learning. Computer Science and Applications 10.6(2020): 10.
4. MLASrivastava, Nitish, et al. Dropout: A Simple Way to Prevent Neural Networks from Overfitting. Journal of Machine Learning Research 15.1(2014):1929-1958.
5. Yao, Yuan, L. Rosasco, and A. Caponnetto. On Early Stopping in Gradient Descent Learning. Constructive Approximation 26.2(2007):289-315.
6. Yang Yinan et al. Research on Small Sample Data Generation Technology Based on Generative Adversarial Network. Electric Power Construction 40.05(2019): 71-77.
7. Zhuang Fuzhen et al. Research Progress on Transfer Learning. Journal of Software 26.1(2015): 26-39.
8. Rajpurkar, P., Park, A., Irvin, J. et al. AppendiXNet: Deep Learning for Diagnosis of Appendicitis from A Small Dataset of CT Exams Using Video Pretraining. Sci Rep 10, 3958 (2020). https://doi.org/10.1038/s41598-020-61055-6
9. J. Deng, W. Dong, R. Socher, L.-J. Li, K. Li, and L. FeiFei. ImageNet: A large-scale hierarchical image database. In IEEE Conference on Computer Vision and Pattern Recognition (CVPR), pages 248–255. IEEE, 2009. 1
10. M. D. Zeiler and R. Fergus. Visualizing and understanding convolutional networks. In European Conference on Computer Vision (ECCV), pages 818–833. Springer, 2014. 1
11. R. Girshick, J. Donahue, T. Darrell, and J. Malik. Rich feature hierarchies for accurate object detection and semantic segmentation. In IEEE Conference on Computer Vision and Pattern Recognition (CVPR), pages 580–587, 2014. 1
12. Yosinski J , Clune J , Bengio Y , et al. How transferable are features in deep neural networks?[C]// International Conference on Neural Information Processing Systems. MIT Press, 2014.
13. A. Torralba, R. Fergus, and W. T. Freeman. 80 million tiny images: A large data set for

nonparametric object and scene recognition. IEEE Transactions on Pattern Analysis and Machine Intelligence (TPAMI), 30(11):1958–1970, Nov 2008. 1
14. C. Zhang, S. Bengio, M. Hardt, B. Recht, and O. Vinyals. Understanding deep learning requires rethinking generalization. In International Conference on Learning Representations (ICLR), 2017. 1
15. B. Barz and J. Denzler, "Deep Learning on Small Datasets without Pre-Training using Cosine Loss," 2020 IEEE Winter Conference on Applications of Computer Vision (WACV), Snowmass, CO, USA, 2020, pp. 1360-1369, doi: 10.1109/WACV45572.2020.9093286.
16. Koppe G , Meyer-Lindenberg A , Durstewitz D . Deep learning for small and big data in psychiatry[J]. Neuropsychopharmacology, 2020:1-17.
17. Krizhevsky, Alex. (2012). Convolutional Deep Belief Networks on CIFAR-10. https://core.ac.uk/display/21813460
18. Nilsback, Maria Elena, and A. Zisserman. Automated Flower Classification over a Large Number of Classes. Sixth Indian Conference on Computer Vision, Graphics & Image Processing, ICVGIP 2008, Bhubaneswar, India, 16-19 December 2008 IEEE , 2008.
19. Zhou Feiyan, Jin Linpeng, and Dong Jun. A Survey of Convolutional Neural Networks. Chinese Journal of Computers 6(2017).
20. He, Kaiming , et al. Deep Residual Learning for Image Recognition[C] IEEE Conference on Computer Vision & Pattern Recognition. IEEE Computer Society, 2016.
21. Huang G , Liu Z , Laurens V , et al. Densely Connected Convolutional Networks[J]. IEEE Computer Society, 2016.
22. Chollet, Francois . Xception: Deep Learning with Depthwise Separable Convolutions. 2017 IEEE Conference on Computer Vision and Pattern Recognition (CVPR) IEEE, 2017.
23. Ioffe, Sergey & Szegedy, Christian. (2015). Batch Normalization: Accelerating Deep Network Training by Reducing Internal Covariate Shift. https://arxiv.org/abs/1502.03167
24. Chieng, Hung Hock, N. Wahid, and P. Ong. Parametric Flatten-T Swish: An Adaptive Nonlinear Activation Function For Deep Learning. Journal of Information and Communication Technology 20.1(2020):21-39.
25. Arkah, Z. M., and L. S. Alzubaidi. Convolutional Neural Network with Global Average Pooling for Image Classification. International Conference on Electrical, Communication, Electronics, Instrumentation and Computing (ICECEIC) 2020.
26. Wan Lei et al. Summary of Application of Softmax Classifier Deep Learning Image Classification Method. Navigation and Control (2019).
27. C. Szegedy, et al. Going deeper with convolutions," in 2015 IEEE Conference on Computer Vision and Pattern Recognition (CVPR), Boston, MA, USA, 2015 pp. 1-9. doi: 10.1109/CVPR .2015.7298594
28. Hamker, Fred H . Predictions of a model of spatial attention using sum- and max-pooling functions. Neurocomputing 56.none(2018):329-343
29. Lin, Min , Q. Chen , and S. Yan . Network In Network. Computer Science (2013).
30. Sandler, Mark  Howard, Andrew  Zhu, Menglong Zhmoginov, Andrey Chen, Liang-Chieh. (2018). Inverted Residuals and Linear Bottlenecks: Mobile Networks for Classification Detection and Segmentation. https://arxiv.org /abs/1801.04381
31.  Simonyan K , Zisserman A . Very Deep Convolutional Networks for Large-Scale Image Recognition[J]. Computer Science, 2014.
32. Szegedy, Christian Ioffe, Sergey Vanhoucke, Vincent Alemi, Alexander. (2016). Inception-v4, Inception-ResNet and the Impact of Residual Connections on Learning. AAAI

Conference on Artificial Intelligence.
33. Tan, Mingxing, and Quoc Le. "Efficientnet: Rethinking model scaling for convolutional neural networks." International Conference on Machine Learning. PMLR, 2019.
34. PyTorch, Training a classifier. [ol].https://pytorch.org/tutorials/beginner/blitz/cifar10 tutorial.html
35. S. M. K. Hasan, Densenet implementation on cifar10 dataset using pytorch. [ol]. https://github.com/SMKamrulHasan/DenseNet-using-PyTorch-CIFAR10.
36. Hasani, Behzad, Pooran Singh Negi, and Mohammad Mahoor. "BReG-NeXt: Facial affect computing using adaptive residual networks with bounded gradient." IEEE Transactions on Affective Computing (2020).
37. Fard, Ali Pourramezan, Hojjat Abdollahi, and Mohammad Mahoor. "ASMNet: A Lightweight Deep Neural Network for Face Alignment and Pose Estimation." Proceedings of the IEEE/CVF Conference on Computer Vision and Pattern Recognition. 2021.
38. Lecun Y, Bottou L. Gradient-based learning applied to document recognition. Proceedings of the IEEE, 1998, 86(11):2278-2324.
39. Xiao H, Rasul K, Vollgraf R. Fashion-MNIST: a Novel Image Dataset for Benchmarking Machine Learning Algorithms. 2017.
40. Vinyals O, Blundell C, Lillicrap T, et al. Matching Networks for One Shot Learning. 2016.
41. Wah C, Branson S, Welinder P, et al. The Caltech-UCSD Birds-200-2011 Dataset. california institute of technology, 2011.